\documentclass{article}

    \PassOptionsToPackage{numbers, compress}{natbib}

\usepackage{multirow}

    \usepackage[preprint]{neurips_2024}

\usepackage[utf8]{inputenc} 
\usepackage[T1]{fontenc}    
\usepackage{hyperref}       
\usepackage{url}            
\usepackage{booktabs}       
\usepackage{amsmath}
\usepackage{amsfonts}       
\usepackage{nicefrac}       
\usepackage{microtype}      
\usepackage{xcolor}         
\usepackage{adjustbox}
\title{Early Exit and Multi-Stage Knowledge Distillation in VLMs for Video Summarization.}

\author{
Anas Anwarul Haq Khan$^{1}$,
Utkarsh Verma$^{2}$,
Ganesh Ramakrishnan$^{1}$ \\
$^{1}$Department of Computer Science and Engineering, IIT Bombay \\
$^{2}$Center of Machine Intelligence and Data Science (C-MInDS), IIT Bombay \\
\texttt{\{anaskhan, ganesh\}@cse.iitb.ac.in}, \texttt{utkverma23@minds.iitb.ac.in}
}

\begin{document}

\maketitle

\begin{abstract}
We introduce \textbf{DEEVISum} (Distilled Early-Exit Vision-language model for Summarization), a lightweight, efficient, and scalable vision-language model designed for segment-wise video summarization. Leveraging multi-modal prompts that combine textual and audio-derived signals, DEEVISum incorporates \textit{Multi-Stage Knowledge Distillation (MSKD)} and \textit{Early Exit (EE)} to strike a balance between performance and efficiency. MSKD offers a 1.33\% absolute F1 improvement over baseline distillation (0.5\%), while EE reduces inference time by approximately 21\% with a 1.3-point drop in F1. Evaluated on the TVSum dataset, our best model—PaLI-Gemma2-3B + MSKD—achieves an F1 score of 61.1, competing the performance of significantly larger models, all while maintaining a lower computational footprint. We publicly release our code and processed dataset\footnote{\url{https://github.com/anas2908/DEEVISum}} to support further research.
\end{abstract}

\section{Introduction}

The digital era has ushered in a shift in video consumption patterns—from traditional long-form content like movies and television shows to brief, attention-grabbing formats popularized by platforms such as TikTok, YouTube Shorts, and Instagram. As audiences increasingly favor content under two minutes, content creators face the challenge of either generating new short videos or condensing existing ones. While manual summarization remains common, it is labor-intensive. 

With the advent of transformer architectures, the field of vision-language modeling has witnessed the rise of billion-parameter models capable of capturing extensive world knowledge. While vision-language models can yield competitive or even superior results compared to traditional approaches that rely heavily on feature engineering, they still suffer from significant limitations—most notably, high computational demands and long inference times. Despite their impressive performance, such large-scale vision-language models (VLMs) are often impractical for real-time or production-level deployment in video summarization tasks due to these constraints. To address these limitations, we propose \textbf{DEEVISum} (\textbf{D}istilled \textbf{E}arly-\textbf{E}xit \textbf{VI}sion-language model for \textbf{Sum}marization), a VLM-based architecture that incorporates multi-stage knowledge distillation and early exits to achieve efficient and scalable video summarization. Knowledge distillation enables the transfer of performance capabilities from large teacher models to smaller student models, significantly reducing computational overhead while maintaining comparable accuracy. In parallel, early exit mechanisms are employed to dynamically determine the optimal depth of computation during inference. Specifically, early exit refers to a technique where intermediate layers of a deep neural network are equipped with auxiliary classifiers, allowing the model to make confident predictions without traversing the entire network. This not only reduces latency but also conserves computational resources by terminating inference as soon as sufficient semantic understanding is achieved. Together, these strategies make DEEVISum well-suited for efficient video summarization in resource-constrained settings.

In this work, we define the problem as follows: given a video $v_i \in \{V_1, V_2, \cdots, V_n\}$, our goal is to use its title ($T_{v_i}$), transcript ($Tr_{v_i}$), and audio-derived annotations such as emotion recognition and speaker diarization ($A_{v_i}$) to identify and retain the most relevant content segments. We approach this task in a supervised setting, focusing on preserving semantic integrity and engagement. In contrast to prior work that centers solely on visual features~\cite{zhang2016video, zhang2018retrospective, zhu2020dsnet, apostolidis2021combining}, our approach leverages pre-trained vision-language models (VLMs), with a strong emphasis on textual and audio-semantic signals.

\noindent Our contributions can be summarized as follows:

\begin{itemize}
    \item \textbf{(1)} We propose a \textbf{multi-stage hierarchical knowledge distillation} strategy that improves performance over standard single-stage distillation methods.
    
    \item \textbf{(2)} We present \textbf{DEEVISum}, a Distilled Early-Exit Vision-Language model that combines large VLM capabilities with computation-aware design, enabling practicality.
    
    \item \textbf{(3)} We design a \textbf{multi-modal supervised framework} that leverages textual cues (titles, transcripts) and audio-derived features (emotion, speaker identity), moving beyond purely visual information.
\end{itemize}

\section{Related Work}

\subsection{Unsupervised Video Summarization}

Unsupervised video summarization techniques \cite{zhou2018deep, de2011vsumm, lee2012discovering, elhamifar2015dissimilarity, jung2019discriminative, zhao2014quasi, he2019unsupervised, jung2020global} do not rely on annotated summaries. Instead, they often employ heuristics such as clustering similar frames to reduce redundancy \cite{gong2000video}. These methods are particularly effective in contexts like egocentric recordings and surveillance footage, where deterministic or handcrafted rules for frame selection can be established. While unsupervised models are appealing for their adaptability and reduced labeling cost, they tend to struggle with high-level semantic understanding and narrative coherence, especially in open-domain content.

\subsection{Supervised Video Summarization}

Supervised methods \cite{zhao2021reconstructive, zhu2020dsnet, fajtl2019summarizing, jiang2022joint, gong2014diverse, narasimhan2021clip, ji2019video, sharghi2017query, zhao2018hsa} learn from manually annotated video summaries \cite{song2015tvsum, gygli2014creating} to predict the importance of frames or segments. Early approaches like DSNet \cite{zhu2020dsnet} rely on deep convolutional features and sequential modeling to exploit temporal dependencies, while VASNet \cite{fajtl2019summarizing} leverages self-attention for keyframe selection, offering better interpretability and lower complexity compared to recurrent networks. PGL-SUM \cite{apostolidis2021combining} further refines this by integrating both local and global attention, improving the model's ability to capture long-range dependencies.

More recent methods explore task-aware and multimodal designs. IPTNet \cite{jiang2022joint} introduces a two-branch architecture that simultaneously handles summarization and moment localization via shared importance maps and cross-task propagation. CLIP-It \cite{narasimhan2021clip} incorporates text guidance through multimodal transformers for generic and query-specific summarization. A2Summ \cite{he2023align} enhances multimodal alignment using contrastive losses and segment-aware embeddings, focusing on better synchronization between textual and visual cues.

\section{Datasets}
\label{sec:dataset}

\begin{table}[htbp]
\centering
\caption{Summary statistics for popular video summarization datasets.}
\begin{adjustbox}{width=\textwidth}
\begin{tabular}{|l|c|c|c|c|c|}
\hline
\textbf{Dataset} & \textbf{Num of Videos} & \textbf{Num of Annotations} & \textbf{Avg Duration} & \textbf{Min Duration} & \textbf{Max Duration} \\
\hline
SumMe~\cite{gygli2014creating}  & 25 & 15--18 & 2min 26s & 32s   & 5min 24s  \\
TVSum~\cite{song2015tvsum}      & 50 & 20     & 3min 55s & 1min 23s & 10min 47s \\
OVP~\cite{li2010multi}          & 50 & 5      & 1min 38s & 46s   & 3min 29s  \\
YouTube~\cite{de2009vsumm}      & 39 & 5      & 3min 16s & 9s    & 9min 32s  \\
\hline
\end{tabular}
\end{adjustbox}
\label{tab:dataset_stats}
\end{table}

For our experiments, we focus on two widely-used video summarization benchmarks: \textbf{SumMe}~\cite{gygli2014creating} and \textbf{TVSum}~\cite{song2015tvsum}. These datasets are standard in the literature and provide well-annotated summaries with diverse content and evaluation protocols.

\textbf{TVSum}~\cite{song2015tvsum} consists of 50 videos sourced from YouTube, spanning 10 categories such as how-to tutorials, news, and documentaries, with 5 videos per category. Each video ranges from 1 to 5 minutes in duration. Frame-level importance scores are collected from 20 users per video by dividing the videos into uniform 2-second shots and aggregating user responses.

\textbf{SumMe}~\cite{gygli2014creating} contains 25 videos covering various real-life scenarios like holidays, cooking, and sports. The videos are typically between 1.5 to 6.5 minutes long. Human annotations are provided in the form of both interval-based and frame-level scores, with at least 15 user summaries per video.

We choose to evaluate exclusively on SumMe and TVSum due to their widespread use in the literature, the availability of multiple human-generated annotations, and the diversity of video content they represent. In addition to these two benchmarks, we reference several auxiliary datasets commonly used in prior works for augmentation and evaluation: \textbf{YouTube}~\cite{de2009vsumm} and \textbf{Open Video Project (OVP)}~\cite{li2010multi} datasets contain 39 and 50 videos respectively. YouTube videos are drawn from categories such as news, sports, and cartoons, while OVP contains a broader range including documentaries and educational content. These datasets provide additional training samples and are typically annotated with interval-based summaries. \textbf{QFVS}~\cite{fu2020attentive} provides ground-truth generic summaries for 4 long-form videos (3–5 hours each) from the \textbf{UT Egocentric}~\cite{lee2012discovering} dataset, captured from head-mounted cameras in uncontrolled environments. The summaries are created via user selection over temporally segmented shots, with final labels computed by averaging annotations from three users.

\section{Proposed Methodology}
\label{sec:methodology}
\begin{figure}[ht]
    \centering
    \includegraphics[width=\linewidth]{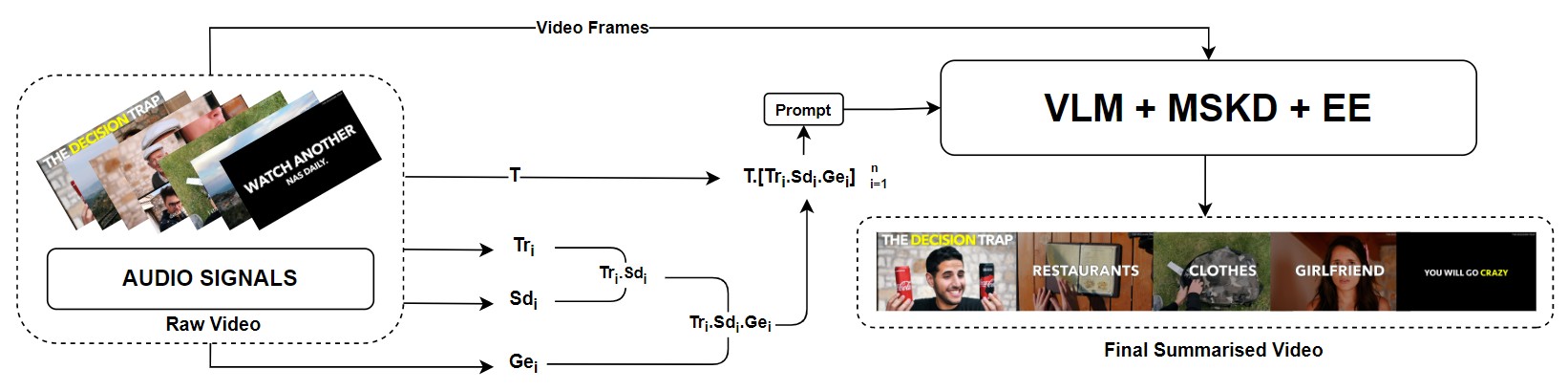}
    \caption{Our final model, \textbf{DEEVISum}, integrates three core innovations: \textit{Early Exit}, \textit{Multi-Stage Knowledge Distillation}, and \textit{Prompt Enhancement through Textual and Audio Information}. $Tr_i$, $Sd_i$, and $Ge_i$ represent the transcript, speaker diarization, and gender-emotion features at the $i$-th second of the video, respectively, while $T$ denotes the video title and $n$ is the length of the video in seconds.}

    \label{fig:vlm_mskd_ee}
\end{figure}

Our goal is to build a vision-language model (VLM) that is both computationally efficient and capable of maintaining high accuracy for video summarization. To this end, we propose two synergistic strategies: \textbf{Multi-Stage Knowledge Distillation (MSKD)} and an \textbf{Early Exit Mechanism in VLMs}. While MSKD ensures effective compression by progressively transferring knowledge across models of decreasing size, the early exit mechanism reduces inference time by allowing confident predictions to be made at intermediate layers. Additionally, we enhance the model’s language input through prompt engineering that integrates audio-derived signals—such as speaker and emotion cues—alongside the video’s title and transcript. Our final model design is illustrated in Figure~\ref{fig:vlm_mskd_ee}.

\subsection{Multi-Stage Knowledge Distillation}
\label{sec:mskd}

Traditional knowledge distillation typically involves a single teacher-student framework. However, this often results in a large knowledge gap between the powerful teacher and a lightweight student, hindering the student’s ability to generalize effectively. 

To address this, we propose a \textbf{Multi-Stage Knowledge Distillation (MSKD)} approach. In MSKD, we introduce an intermediate-sized \textbf{mentor} model between the teacher and the student. This facilitates a smoother and more structured transfer of knowledge across models, reducing the distillation difficulty at each stage.

In our setting:
\begin{itemize}
    \item The \textbf{Teacher} is the large model (e.g. Paligemma2-28B),
    \item The \textbf{Mentor} is an intermediate model (e.g. Paligemma2-10B),
    \item The \textbf{Student} is a smaller, efficient model (e.g. Paligemma2-3B).
\end{itemize}

\begin{figure}[ht]
    \centering
    \includegraphics[width=0.9\linewidth]{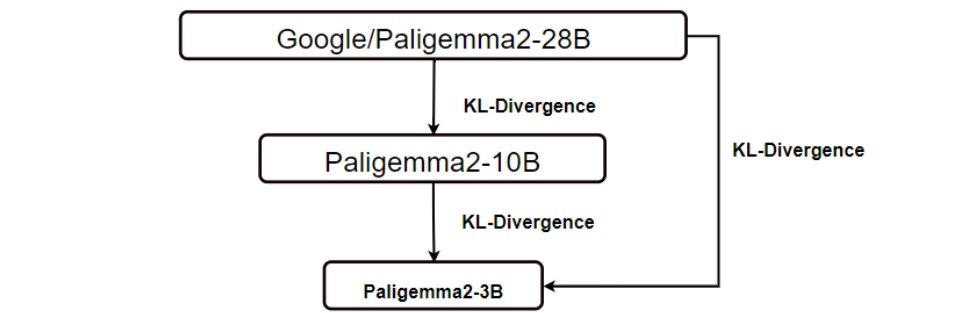}
    \caption{Proposed Multi-Stage Knowledge Distillation (MSKD) pipeline with teacher, mentor, and student models using KL-divergence-based loss at each stage.}
    \label{fig:mskd}
\end{figure}

The MSKD pipeline is illustrated in Figure~\ref{fig:mskd}. The objective functions at each stage are defined as follows:

\paragraph{Teacher Loss:}
\begin{equation}
\mathcal{L}_{\text{teacher}} = \mathcal{L}_{\text{CE}}(y, \hat{y})
\end{equation}
Here, the teacher is trained using standard cross-entropy with the gold labels.

\paragraph{Mentor Loss:}
\begin{equation}
\mathcal{L}_{\text{mentor}} = \mathcal{L}_{\text{CE}}(y, \hat{y}_m) + \phi \cdot D_{\text{KL}}(P_m \parallel P_t)
\end{equation}
The mentor minimizes both cross-entropy and KL divergence with the teacher’s distribution, guided by a balancing factor $\phi$.

\paragraph{Student Loss:}
\begin{equation}
\mathcal{L}_{\text{student}} = \mathcal{L}_{\text{CE}}(y, \hat{y}_s) + \phi \cdot D_{\text{KL}}(P_s \parallel P_m) + \psi \cdot D_{\text{KL}}(P_s \parallel P_t)
\end{equation}
The student is encouraged to learn from both the mentor and the teacher simultaneously, using separate KL divergence terms, balanced by weights $\phi$ and $\psi$.

This multi-stage framework offers a gradual and stable learning path for the student and is particularly suited for VLMs, where textual and visual modalities can lead to complex feature interactions that benefit from intermediate supervision.

\subsection{MuE-style Early Exit via Cosine-Similarity Confidence}
\label{sec:adaptive_early_exit}

We implement a dynamic early-exit strategy inspired by the MuE model~\cite{li2023you} to accelerate inference in our vision-language summarization pipeline. The core idea is to bypass full-depth decoding for semantically simple inputs by enabling confident predictions at intermediate decoder stages. Unlike the original MuE, where intermediate exits are trained jointly, our approach applies early exit only during inference, with no modification to training procedures.

\paragraph{Exit Architecture.} We augment the decoder stack of the student vision-language model with $N$ exit heads placed at fixed depths $\{d_1, d_2, \dots, d_N\}$. Each exit head consists of a lightweight decoder module followed by a classification layer that predicts the binary summarization label. These heads reuse the encoder output and decode using fewer transformer blocks, enabling fast local decisions for frame or segment-level predictions.

\paragraph{Cosine Similarity-Based Confidence Scoring.} To decide whether an input can exit early, we compute a confidence score for each exit head based on the cosine similarity between its prediction vector $\mathbf{p}_i$ and a learned summary-class prototype vector $\mathbf{q}$:
\begin{equation}
\gamma_i = \frac{\mathbf{p}_i^\top \mathbf{q}}{\|\mathbf{p}_i\| \cdot \|\mathbf{q}\|}
\end{equation}
This formulation captures the semantic alignment between the current prediction and the summary class representation. The prototype vector $\mathbf{q}$ is learned jointly with the main model and stored for inference.

\paragraph{Early Exit Routing Policy.} During inference, each input is processed sequentially through the decoder exits. At the $i$-th exit, we compute $\gamma_i$. If $\gamma_i \geq \tau$, where $\tau \in (0,1)$ is a fixed threshold, we terminate computation and return the corresponding prediction $\hat{y}_i$. Otherwise, the input proceeds to the next exit:
\begin{equation}
\text{Output}(x) =
\begin{cases}
\hat{y}_i, & \text{if } \gamma_i \geq \tau \\ \hat{y}_N, & \text{otherwise}
\end{cases}
\end{equation}
This gating mechanism dynamically adjusts the compute path per input, reducing overall decoding cost without compromising outputs on complex samples.

\paragraph{Training Protocol.} To preserve training simplicity, no supervision is applied to intermediate exit heads. The model is trained using a standard cross-entropy objective at the final decoder. The early-exit policy is activated only during inference, which avoids interference with optimization dynamics and requires no changes to the base model’s loss or scheduling.

\begin{figure}[htbp]
    \centering
    \includegraphics[width=\linewidth]{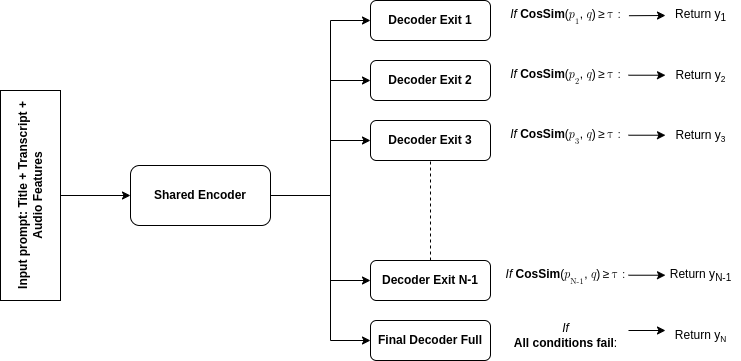}  
    \caption{Illustration of the early exit mechanism used during inference. Exit heads compute cosine similarity with a learned prototype vector, and if confidence exceeds threshold $\tau$, the prediction is returned early. Otherwise, the sample continues to deeper exits until reaching the final decoder.}
    \label{fig:early_exit_mue}
\end{figure}

\paragraph{Design Rationale and Efficiency Trade-off.} This inference-only variant of MuE provides the benefits of dynamic computation without retraining. By leveraging the semantic alignment signal captured via cosine similarity, we minimize average latency across a dataset by skipping deeper decoding for high-confidence segments. The value of $\tau$ is tuned empirically on the validation set to balance accuracy and inference time. Figure~\ref{fig:early_exit_mue} illustrates the full early exit mechanism including confidence routing and fallback logic.

\newpage

\subsection{Audio Processing}
\label{subsec:audio_processing}

\textbf{Audio Processing:} To enrich the semantic understanding of videos beyond visual cues, we incorporate audio-based signals—specifically speaker diarization and gender-emotion classification. These cues offer valuable context, especially in scenarios where speakers are not visible or where audio conveys critical emotional tone.

\begin{itemize}
    \item \textbf{Speaker Diarization:} We employ the \textit{pyannote.audio} toolkit~\cite{bredin2020pyannote} to detect and separate distinct speakers in the audio stream. This allows us to segment the video based on speaker turns and interactions, which is especially useful for analyzing dialogues and discussions.

    \item \textbf{Gender-Emotion Classification:} Relying solely on transcripts may lead to the loss of affective information, particularly in videos with limited visual cues (e.g., voice-over narrations, commentary). To address this, we segment the audio into 3-second windows and use a deep learning model\footnote{\url{https://github.com/MiteshPuthran/Speech-Emotion-Analyzer}} trained on four publicly available datasets—SAVEE~\cite{jackson2014surrey}, RAVDESS~\cite{livingstone2018ryerson}, TESS~\cite{dupuis2010toronto}, and CREMA-D~\cite{cao2014crema}—to classify both the speaker’s emotion and gender. This information enriches the representation of audio segments and provides a nuanced understanding of speaker affect and identity.
\end{itemize}

The extracted audio-based annotations—speaker diarization and gender-emotion cues—are incorporated alongside the video title ($T_{v_i}$) and transcript ($Tr_{v_i}$) to form a comprehensive language input for the vision-language model. Let $A_{v_i}$ denote the aggregated audio features for video $v_i$. The final textual input to the language encoder of the VLM is structured as a prompt-conditioned sequence combining $T_{v_i}$, $Tr_{v_i}$, and $A_{v_i}$, enabling the model to reason over rich, multi-source semantic content.

\section{Evaluation Metric} \label{sec:evaluation_metrics}

For automatic evaluation, we adopt the F1-score metric, widely used in prior works, to assess the overlap between the predicted summary and the ground truth at the frame level. Let $G$ and $S$ denote the sets of selected frames in the ground truth and system-generated summaries, respectively. Then, Precision ($P$), Recall ($R$), and F1-score are defined as:

\begin{equation}
P = \frac{|G \cap S|}{|S|}, \quad R = \frac{|G \cap S|}{|G|}, \quad F1 = \frac{2PR}{P + R}
\end{equation}

\section{Experiments and Results}

\subsection{SoTA Models}

\begin{table}[htbp]
\centering
\caption{Comparison of F1-scores on SumMe and TVSum datasets across selected baselines.}
\label{tab:f1_baseline_comparison}
\begin{tabular}{|l|c|c|c|c|c|}
\hline
\textbf{Dataset} & \textbf{Random} & \textbf{VASNet} & \textbf{DSNet} & \textbf{iPTNet} & \textbf{A2Summ} \\
\hline
\textbf{SumMe (F1)} & 41.0 & 49.7 & 51.2 & 54.5 & \textbf{55.0} \\
\textbf{TVSum (F1)} & 57.0 & 61.4 & 61.9 & \textbf{63.4} & \textbf{63.4} \\
\hline
\end{tabular}
\end{table}

Table~\ref{tab:f1_baseline_comparison} presents a comparative analysis of F1-scores for selected video summarization baselines on the SumMe and TVSum datasets. As expected, the Random baseline performs the worst, with scores of 41.0 on SumMe and 57.0 on TVSum, setting a lower bound for performance. Among learning-based methods, A2Summ leads with the highest F1-score on SumMe (55.0), slightly outperforming iPTNet (54.5) and DSNet-AF (51.2). On the TVSum dataset, both iPTNet and A2Summ achieve the best F1-score of 63.4. VASNet, despite being one of the earlier approaches, remains competitive with 49.7 on SumMe and 61.4 on TVSum.

\subsection{Prompt Enhancement through Meta Information}

\noindent
To better understand the contribution of different input modalities in guiding a vision-language model (VLM) for frame-level importance estimation, we perform an ablation study using the PaLI-Gemma-3B model without applying any Knowledge Distillation or Early Exit mechanisms. The model is prompted using the following instruction: \textit{``Given a video and its <Meta Information> (title, transcript, speaker and emotion cues), assign each frame a score from 1 to 5 based on its semantic similarity to the meta information—5 for highest relevance, 1 for lowest in context of Video Summarization.''} We progressively include textual and audio cues in the input prompt to observe their effect on summarization performance. Here, \textbf{T} denotes the video title, \textbf{Tr} is the transcript, \textbf{Ge} represents Gender-Emotion cues, and \textbf{Sd} indicates Speaker Diarization.

\begin{table}[htbp]
\centering
\caption{F1 scores on TVSum for different combinations of textual and audio inputs.}
\label{tab:f1_component_ablation}
\begin{tabular}{|c|c|c|c|c|}
\hline
\textbf{Input Modality} & \textbf{T} & \textbf{T + Tr} & \textbf{T + Tr + Ge} & \textbf{T + Tr + Ge + Sd} \\
\hline
\textbf{F1 Score} & 59.2 & 60.4 & 60.7 & 60.3 \\
\hline
\end{tabular}
\end{table}

\noindent
The ablation results in Table~\ref{tab:f1_component_ablation} reveal clear trends in how different components influence performance. Using only the title (\textbf{T}) yields a modest F1 score of 59.2, indicating that high-level video metadata alone is insufficient for detailed summarization. When transcripts (\textbf{Tr}) are included, the F1 score significantly increases to 60.4, demonstrating that fine-grained temporal descriptions improve the model’s ability to localize important content. Adding gender-emotion cues (\textbf{Ge}) further boosts performance to 60.7, validating the hypothesis that affective audio features enrich semantic understanding, especially in emotionally expressive or speech-driven content. However, the inclusion of speaker diarization (\textbf{Sd}) leads to a slight decrease to 60.3. We hypothesize this degradation is due to two factors: (1) segmentation noise—diarization errors such as incorrect speaker boundaries may introduce ambiguity in the input prompt, and (2) speaker identity may not always correlate with content importance, especially in videos where the speaker's presence is uniform or does not change the semantic relevance of frames.

\subsection{Scaling PaLI-Gemma-2: 3B, 10B, and 28B Models}
\label{subsec:paligemma_scaling}

To assess the impact of model scale on video summarization performance, we fine-tuned three variants of the PaLI-Gemma-2 vision-language model—3B, 10B, and 28B parameters—on the TVSum dataset. All models were evaluated without the use of knowledge distillation or early exit mechanisms to isolate the effect of model capacity.

\begin{table}[htbp]
\centering
\caption{F1 scores for different sizes of PaLI-Gemma-2 models on the TVSum dataset.}
\label{tab:f1_model_scaling}
\begin{tabular}{|c|c|c|c|}
\hline
\textbf{Model} & \textbf{PaLI-Gemma2-3B} & \textbf{PaLI-Gemma2-10B} & \textbf{PaLI-Gemma2-28B} \\
\hline
\textbf{F1 Score} & 60.3 & 61.6 & 62.2 \\
\hline
\end{tabular}
\end{table}

As shown in Table~\ref{tab:f1_model_scaling}, performance improves consistently with model scale, suggesting that larger models are better able to capture the semantic relationships between visual frames and accompanying textual and audio metadata. The 3B model achieves an F1 score of 60.3, while the 10B and 28B models attain 61.6 and 62.4 respectively, reflecting gains of 1.3 and 2.1 points over the smallest variant. We hypothesize that the improvements stem from the increased representational capacity of larger models, which are more adept at integrating multimodal information and learning fine-grained associations. However, these performance gains come at the cost of increased computational overhead, which may limit their applicability in latency-sensitive or resource-constrained environments.

\subsection{Knowledge Distillation and Multi-Stage Distillation for PaLI-Gemma-2 Models}
\label{subsec:kd_mskd_analysis}

We evaluate the impact of Knowledge Distillation (KD) and Multi-Stage Knowledge Distillation (MSKD) on video summarization using different configurations of PaLI-Gemma-2 vision-language models with 3B, 10B, and 28B parameters. The largest model (28B) is treated as the teacher in all distillation settings.

\begin{table}[htbp]
\centering
\caption{F1 scores on TVSum with and without knowledge distillation (KD), and percentage improvement.}
\label{tab:kd_comparison}
\begin{adjustbox}{max width=\textwidth}
\begin{tabular}{|c|c|c|c|c|c|}
\hline
\textbf{Model Variant} & \textbf{VLM 28B} & \textbf{VLM 10B} & \textbf{VLM 3B (10B KD)} & \textbf{VLM 3B (28B KD)} & \textbf{VLM 3B (MSKD)} \\
\hline
\textbf{F1 Score (No KD)} & 62.2 & 61.6 & 60.3 & 60.3 & 60.3 \\
\hline
\textbf{F1 Score (With KD)} & 62.2 & 61.9 & 60.9 & 60.6 & 61.1 \\
\hline
\textbf{\% Improvement} & 0.0\% & +0.49\% & +1.00\% & +0.50\% & +1.33\% \\
\hline
\end{tabular}
\end{adjustbox}
\end{table}

From Table~\ref{tab:kd_comparison}, we observe that incorporating knowledge distillation consistently improves the performance across all model sizes. The 3B model improves from 60.3 to 60.9 and 60.6 when distilled from 10B and 28B models, respectively. The best performance for the 3B model is observed with the Multi-Stage Knowledge Distillation (MSKD) approach, achieving an F1 score of 61.1.

\textbf{Model Distillation Setups and Loss Functions:}
\begin{itemize}
    \item \textbf{VLM 28B:} Acts as the teacher model; no distillation is applied, and the model is trained using standard cross-entropy loss $\mathcal{L}_{\text{CE}}(y, \hat{y})$.

    \item \textbf{VLM 10B (28B as teacher):}
    \begin{equation}
        \mathcal{L}_{\text{student}} = \mathcal{L}_{\text{CE}}(y, \hat{y}_s) + \lambda \cdot D_{\text{KL}}(P_{10B} \parallel P_{28B})
    \end{equation}
    \item \textbf{VLM 3B (10B as teacher):}
    \begin{equation}
        \mathcal{L}_{\text{student}} = \mathcal{L}_{\text{CE}}(y, \hat{y}_s) + \lambda \cdot D_{\text{KL}}(P_{3B} \parallel P_{10B})
    \end{equation}
    \item \textbf{VLM 3B (28B as teacher):}
    \begin{equation}
        \mathcal{L}_{\text{student}} = \mathcal{L}_{\text{CE}}(y, \hat{y}_s) + \lambda \cdot D_{\text{KL}}(P_{3B} \parallel P_{28B})
    \end{equation}
    \item \textbf{VLM 3B (MSKD: 28B teacher, 10B mentor):}
    \begin{equation}
        \mathcal{L}_{\text{student}} = \mathcal{L}_{\text{CE}}(y, \hat{y}_s) + \phi \cdot D_{\text{KL}}(P_{3B} \parallel P_{10B}) + \psi \cdot D_{\text{KL}}(P_{3B} \parallel P_{28B})
    \end{equation}
\end{itemize}

\textbf{Analysis and Hypothesis:}  
The inclusion of a hierarchical distillation structure in \textbf{Multi-Stage Knowledge Distillation (MSKD)} leads to the best performance among all evaluated strategies, as seen in Table~\ref{tab:kd_comparison}. This improvement can be attributed to the presence of an intermediate \textbf{mentor model} (PaLI-Gemma-2 10B), which effectively bridges the representational and capacity gap between the large \textbf{teacher model} (28B) and the small \textbf{student model} (3B). The mentor assists in gradually transferring knowledge, reducing the difficulty of direct compression and enabling the student to generalize better.

Crucially, during MSKD, the mentor model is \emph{simultaneously trained} with supervision from the teacher, while the student model is jointly trained with supervision from both the mentor and the teacher. This concurrent optimization helps maintain alignment across the model hierarchy and prevents representational drift. We use KL-divergence as the distillation objective and set the hyperparameters $\phi = 0.5$ and $\psi = 0.25$ to balance the influence of the mentor and teacher distributions in the student’s training.

The full set of loss functions used in MSKD across the three models per iteration are as follows:

\begin{equation}
\mathcal{L}_{\text{teacher}} = \mathcal{L}_{\text{CE}}(y, \hat{y})
\end{equation}

\begin{equation}
\mathcal{L}_{\text{mentor}} = \mathcal{L}_{\text{CE}}(y, \hat{y}_m) + \phi \cdot D_{\text{KL}}(P_m \parallel P_t)
\end{equation}

\begin{equation}
\mathcal{L}_{\text{student}} = \mathcal{L}_{\text{CE}}(y, \hat{y}_s) + \phi \cdot D_{\text{KL}}(P_s \parallel P_m) + \psi \cdot D_{\text{KL}}(P_s \parallel P_t)
\end{equation}

Interestingly, we observe that directly distilling from the 28B teacher to the 3B student results in lower performance (F1: 60.6) than using the 10B model as the teacher (F1: 60.9). This supports the hypothesis that very large teachers may produce complex output distributions that are too difficult for smaller students to approximate effectively. By introducing a mentor, MSKD provides a more interpretable and learnable intermediary signal, yielding a better balance between compression and performance in vision-language tasks like video summarization.

\subsection{Impact of Early Exit on Inference Time and Accuracy}

We evaluate the trade-off between inference time and accuracy by comparing the \textbf{PaLI-Gemma2-3B + MSKD} model with its variant augmented by an \textbf{Early Exit (EE)} mechanism. Results are averaged over 15 test videos, each with 20 human-annotated ground truth summaries—a standard practice followed throughout our evaluations.

\begin{table}[htbp]
\centering
\caption{Effect of Early Exit on F1 Score and Inference Time (averaged over 15 videos).}
\label{tab:ee_impact}
\begin{tabular}{|c|c|c|}
\hline
\textbf{Metric} & \textbf{PaLI-Gemma2-3B + MSKD} & \textbf{PaLI-Gemma2-3B + MSKD + EE} \\
\hline
\textbf{F1 Score} & 61.1 & 59.8 \\
\hline
\textbf{Inference Time (s)} & $\approx 715$ sec & $\approx 507$ sec \\
\hline
\end{tabular}
\end{table}

Early Exit achieves a $ \approx\textbf{~21\% reduction} $ in inference time, demonstrating its effectiveness in accelerating model response. However, this comes at the cost of a \textbf{~2.1\% absolute drop in F1 score}, which is a significant degradation compared to other configurations. We hypothesize that this drop stems from architectural limitations of the 3B model, which was not extensively optimized for early exit during inference. Further architecture-aware tuning may mitigate this performance loss in future.

\section{Conclusion}
In this work, we demonstrate the effectiveness of vision-language models (VLMs), particularly Google's PaLI-Gemma family, for video summarization. We fine-tune 3B, 10B, and 28B parameter models using prompts enriched with textual (title, transcript) and audio-derived (speaker diarization, gender-emotion) metadata to enhance multimodal understanding. Performance improves with model scale and is further enhanced through knowledge distillation (KD), with a modest 0.5\% F1 gain. However, introducing multi-stage knowledge distillation (MSKD) significantly boosts performance by 1.33\%, highlighting the benefit of hierarchical supervision from both a 10B mentor and 28B teacher. Early-exit (EE) strategies additionally reduce inference time by approximately 21\%, making summarization more efficient with compromise in accuracy.

\section{Limitations}

Most existing video summarization datasets such as TVSum and SumMe consist of short, templated videos with simplistic annotation schemes. These benchmarks fall short in evaluating models for real-world summarization tasks that demand deeper multimodal understanding and complex reasoning. To highlight this, we evaluated prior models on a curated set of 50 manually annotated videos (4–8 minutes each) across diverse genres including drama, sports, politics, and entertainment. As shown in Table~\ref{tab:results_curated}, prior models perform poorly, while our VLM-based method significantly outperforms them.

\begin{table}[htbp]
\centering
\caption{F1-scores of prior models and our VLM-based model on a curated multimodal dataset.}
\label{tab:results_curated}
\begin{tabular}{|l|c|}
\hline
\textbf{Model} & \textbf{F1-Score} \\
\hline
Random \cite{otani2019rethinking}       & 11.28 \\
DSNet \cite{zhu2020dsnet}               & 17.01 \\
PGL-SUM \cite{apostolidis2021combining} & 17.80 \\
VASNet \cite{fajtl2019summarizing}      & 19.30 \\
IPTNet \cite{jiang2022joint}            & 19.47 \\
A2Summ \cite{he2023align}               & 23.96 \\
\textbf{Ours (PaLI-Gemma2-3B + MSKD + EE)} & \textbf{41.12} \\
\hline
\end{tabular}
\end{table}

These findings underscore that the primary bottleneck in advancing video summarization may no longer be architecture alone, but the limited diversity and depth of current benchmarks. This motivates the need for richer, semantically annotated, and genre-diverse datasets to truly measure progress in multimodal summarization.

\bibliographystyle{plainnat}

\begin{thebibliography}{33}
\providecommand{\natexlab}[1]{#1}
\providecommand{\url}[1]{\texttt{#1}}
\expandafter\ifx\csname urlstyle\endcsname\relax
  \providecommand{\doi}[1]{doi: #1}\else
  \providecommand{\doi}{doi: \begingroup \urlstyle{rm}\Url}\fi

\bibitem[Apostolidis et~al.(2021)Apostolidis, Balaouras, Mezaris, and Patras]{apostolidis2021combining}
Evlampios Apostolidis, Georgios Balaouras, Vasileios Mezaris, and Ioannis Patras.
\newblock Combining global and local attention with positional encoding for video summarization.
\newblock In \emph{2021 IEEE international symposium on multimedia (ISM)}, pages 226--234. IEEE, 2021.

\bibitem[Bredin et~al.(2020)Bredin, Yin, Coria, Gelly, Korshunov, Lavechin, Fustes, Titeux, Bouaziz, and Gill]{bredin2020pyannote}
Herv{\'e} Bredin, Ruiqing Yin, Juan~Manuel Coria, Gregory Gelly, Pavel Korshunov, Marvin Lavechin, Diego Fustes, Hadrien Titeux, Wassim Bouaziz, and Marie-Philippe Gill.
\newblock Pyannote. audio: neural building blocks for speaker diarization.
\newblock In \emph{ICASSP 2020-2020 IEEE International Conference on Acoustics, Speech and Signal Processing (ICASSP)}, pages 7124--7128. IEEE, 2020.

\bibitem[De~Avila et~al.(2011{\natexlab{a}})De~Avila, Lopes, da~Luz~Jr, and de~Albuquerque~Ara{\'u}jo]{de2011vsumm}
Sandra Eliza~Fontes De~Avila, Ana Paula~Brandao Lopes, Antonio da~Luz~Jr, and Arnaldo de~Albuquerque~Ara{\'u}jo.
\newblock Vsumm: A mechanism designed to produce static video summaries and a novel evaluation method.
\newblock \emph{Pattern recognition letters}, 32\penalty0 (1):\penalty0 56--68, 2011{\natexlab{a}}.

\bibitem[De~Avila et~al.(2011{\natexlab{b}})De~Avila, Lopes, da~Luz~Jr, and de~Albuquerque~Araujo]{de2009vsumm}
Sandra Eliza~Fontes De~Avila, Ana Paula~Brandao Lopes, Antonio~da da~Luz~Jr, and Arnaldo de~Albuquerque~Araujo.
\newblock Vsumm: A mechanism designed to produce static video summaries and a novel evaluation method.
\newblock In \emph{Pattern Recognition Letters}, volume~32, pages 56--68. Elsevier, 2011{\natexlab{b}}.

\bibitem[Dupuis and Pichora-Fuller(2010)]{dupuis2010toronto}
Kate Dupuis and M~Kathleen Pichora-Fuller.
\newblock \emph{Toronto emotional speech set (TESS)}.
\newblock University of Toronto, Psychology Department, 2010.

\bibitem[Elhamifar et~al.(2015)Elhamifar, Sapiro, and Sastry]{elhamifar2015dissimilarity}
Ehsan Elhamifar, Guillermo Sapiro, and S~Shankar Sastry.
\newblock Dissimilarity-based sparse subset selection.
\newblock \emph{IEEE transactions on pattern analysis and machine intelligence}, 38\penalty0 (11):\penalty0 2182--2197, 2015.

\bibitem[Fajtl et~al.(2019)Fajtl, Sokeh, Argyriou, Monekosso, and Remagnino]{fajtl2019summarizing}
Jiri Fajtl, Hajar~Sadeghi Sokeh, Vasileios Argyriou, Dorothy Monekosso, and Paolo Remagnino.
\newblock Summarizing videos with attention.
\newblock In \emph{Computer Vision--ACCV 2018 Workshops: 14th Asian Conference on Computer Vision, Perth, Australia, December 2--6, 2018, Revised Selected Papers 14}, pages 39--54. Springer, 2019.

\bibitem[Fu et~al.(2020)Fu, Xu, Tao, and Xu]{fu2020attentive}
Yongqing Fu, Mingfei Xu, Dacheng Tao, and Changsheng Xu.
\newblock Attentive and adversarial learning for video summarization.
\newblock In \emph{AAAI Conference on Artificial Intelligence}, volume~34, pages 10861--10868, 2020.

\bibitem[Gong et~al.(2014)Gong, Chao, Grauman, and Sha]{gong2014diverse}
Boqing Gong, Wei-Lun Chao, Kristen Grauman, and Fei Sha.
\newblock Diverse sequential subset selection for supervised video summarization.
\newblock \emph{Advances in neural information processing systems}, 27, 2014.

\bibitem[Gong and Liu(2000)]{gong2000video}
Yihong Gong and Xin Liu.
\newblock Video summarization using singular value decomposition.
\newblock In \emph{Proceedings IEEE Conference on Computer Vision and Pattern Recognition. CVPR 2000 (Cat. No. PR00662)}, volume~2, pages 174--180. IEEE, 2000.

\bibitem[Gygli et~al.(2014)Gygli, Grabner, Riemenschneider, and Van~Gool]{gygli2014creating}
Michael Gygli, Helmut Grabner, Hayko Riemenschneider, and Luc Van~Gool.
\newblock Creating summaries from user videos.
\newblock In \emph{Computer Vision--ECCV 2014: 13th European Conference, Zurich, Switzerland, September 6-12, 2014, Proceedings, Part VII 13}, pages 505--520. Springer, 2014.

\bibitem[He et~al.(2023)He, Wang, Qiu, Bui, Shrivastava, and Wang]{he2023align}
Bo~He, Jun Wang, Jielin Qiu, Trung Bui, Abhinav Shrivastava, and Zhaowen Wang.
\newblock Align and attend: Multimodal summarization with dual contrastive losses.
\newblock In \emph{Proceedings of the IEEE/CVF conference on computer vision and pattern recognition}, pages 14867--14878, 2023.

\bibitem[He et~al.(2019)He, Hua, Song, Zhang, Xue, Ma, Robertson, and Guan]{he2019unsupervised}
Xufeng He, Yang Hua, Tao Song, Zongpu Zhang, Zhengui Xue, Ruhui Ma, Neil Robertson, and Haibing Guan.
\newblock Unsupervised video summarization with attentive conditional generative adversarial networks.
\newblock In \emph{Proceedings of the 27th ACM International Conference on multimedia}, pages 2296--2304, 2019.

\bibitem[Jackson and Haq(2014)]{jackson2014surrey}
Philip Jackson and SJUoSG Haq.
\newblock Surrey audio-visual expressed emotion (savee) database.
\newblock \emph{University of Surrey: Guildford, UK}, 2014.

\bibitem[Ji et~al.(2019)Ji, Xiong, Pang, and Li]{ji2019video}
Zhong Ji, Kailin Xiong, Yanwei Pang, and Xuelong Li.
\newblock Video summarization with attention-based encoder--decoder networks.
\newblock \emph{IEEE Transactions on Circuits and Systems for Video Technology}, 30\penalty0 (6):\penalty0 1709--1717, 2019.

\bibitem[Jiang and Mu(2022)]{jiang2022joint}
Hao Jiang and Yadong Mu.
\newblock Joint video summarization and moment localization by cross-task sample transfer.
\newblock In \emph{Proceedings of the IEEE/CVF Conference on Computer Vision and Pattern Recognition}, pages 16388--16398, 2022.

\bibitem[Jung et~al.(2019)Jung, Cho, Kim, Woo, and Kweon]{jung2019discriminative}
Yunjae Jung, Donghyeon Cho, Dahun Kim, Sanghyun Woo, and In~So Kweon.
\newblock Discriminative feature learning for unsupervised video summarization.
\newblock In \emph{Proceedings of the AAAI Conference on artificial intelligence}, volume~33, pages 8537--8544, 2019.

\bibitem[Jung et~al.(2020)Jung, Cho, Woo, and Kweon]{jung2020global}
Yunjae Jung, Donghyeon Cho, Sanghyun Woo, and In~So Kweon.
\newblock Global-and-local relative position embedding for unsupervised video summarization.
\newblock In \emph{European Conference on Computer Vision}, pages 167--183. Springer, 2020.

\bibitem[Lee et~al.(2012)Lee, Ghosh, and Grauman]{lee2012discovering}
Yong~Jae Lee, Joydeep Ghosh, and Kristen Grauman.
\newblock Discovering important people and objects for egocentric video summarization.
\newblock In \emph{2012 IEEE conference on computer vision and pattern recognition}, pages 1346--1353. IEEE, 2012.

\bibitem[Li et~al.(2023)Li, Qi, and Lin]{li2023you}
Yunfan Li, Xiaojuan Qi, and Dahua Lin.
\newblock You need multiple exiting: Dynamic early exiting for accelerating unified vision language model.
\newblock In \emph{Proceedings of the IEEE/CVF Conference on Computer Vision and Pattern Recognition (CVPR)}, pages 17189--17200, 2023.

\bibitem[Li et~al.(2010)Li, Ishwar, and Konrad]{li2010multi}
Zhu Li, Prakash Ishwar, and Janusz Konrad.
\newblock Multi-video summarization based on video-mmr.
\newblock In \emph{International Conference on Image Processing (ICIP)}, pages 1477--1480. IEEE, 2010.

\bibitem[Livingstone and Russo(2018)]{livingstone2018ryerson}
Steven~R Livingstone and Frank~A Russo.
\newblock The ryerson audio-visual database of emotional speech and song (ravdess): A dynamic, multimodal set of facial and vocal expressions in north american english.
\newblock \emph{PloS one}, 13\penalty0 (5):\penalty0 e0196391, 2018.

\bibitem[Narasimhan et~al.(2021)Narasimhan, Rohrbach, and Darrell]{narasimhan2021clip}
Medhini Narasimhan, Anna Rohrbach, and Trevor Darrell.
\newblock Clip-it! language-guided video summarization.
\newblock \emph{Advances in neural information processing systems}, 34:\penalty0 13988--14000, 2021.

\bibitem[Otani et~al.(2019)Otani, Nakashima, Rahtu, and Heikkila]{otani2019rethinking}
Mayu Otani, Yuta Nakashima, Esa Rahtu, and Janne Heikkila.
\newblock Rethinking the evaluation of video summaries.
\newblock In \emph{Proceedings of the IEEE/CVF conference on computer vision and pattern recognition}, pages 7596--7604, 2019.

\bibitem[Sharghi et~al.(2017)Sharghi, Laurel, and Gong]{sharghi2017query}
Aidean Sharghi, Jacob~S Laurel, and Boqing Gong.
\newblock Query-focused video summarization: Dataset, evaluation, and a memory network based approach.
\newblock In \emph{Proceedings of the IEEE conference on computer vision and pattern recognition}, pages 4788--4797, 2017.

\bibitem[Song et~al.(2015)Song, Vallmitjana, Stent, and Jaimes]{song2015tvsum}
Yale Song, Jordi Vallmitjana, Amanda Stent, and Alejandro Jaimes.
\newblock Tvsum: Summarizing web videos using titles.
\newblock In \emph{Proceedings of the IEEE conference on computer vision and pattern recognition}, pages 5179--5187, 2015.

\bibitem[Zhang et~al.(2016)Zhang, Chao, Sha, and Grauman]{zhang2016video}
Ke~Zhang, Wei-Lun Chao, Fei Sha, and Kristen Grauman.
\newblock Video summarization with long short-term memory.
\newblock In \emph{Computer Vision--ECCV 2016: 14th European Conference, Amsterdam, The Netherlands, October 11--14, 2016, Proceedings, Part VII 14}, pages 766--782. Springer, 2016.

\bibitem[Zhang et~al.(2018)Zhang, Grauman, and Sha]{zhang2018retrospective}
Ke~Zhang, Kristen Grauman, and Fei Sha.
\newblock Retrospective encoders for video summarization.
\newblock In \emph{Proceedings of the European conference on computer vision (ECCV)}, pages 383--399, 2018.

\bibitem[Zhao and Xing(2014)]{zhao2014quasi}
Bin Zhao and Eric~P Xing.
\newblock Quasi real-time summarization for consumer videos.
\newblock In \emph{Proceedings of the IEEE conference on computer vision and pattern recognition}, pages 2513--2520, 2014.

\bibitem[Zhao et~al.(2018)Zhao, Li, and Lu]{zhao2018hsa}
Bin Zhao, Xuelong Li, and Xiaoqiang Lu.
\newblock Hsa-rnn: Hierarchical structure-adaptive rnn for video summarization.
\newblock In \emph{Proceedings of the IEEE conference on computer vision and pattern recognition}, pages 7405--7414, 2018.

\bibitem[Zhao et~al.(2021)Zhao, Li, Lu, and Li]{zhao2021reconstructive}
Bin Zhao, Haopeng Li, Xiaoqiang Lu, and Xuelong Li.
\newblock Reconstructive sequence-graph network for video summarization.
\newblock \emph{IEEE Transactions on Pattern Analysis and Machine Intelligence}, 44\penalty0 (5):\penalty0 2793--2801, 2021.

\bibitem[Zhou et~al.(2018)Zhou, Qiao, and Xiang]{zhou2018deep}
Kaiyang Zhou, Yu~Qiao, and Tao Xiang.
\newblock Deep reinforcement learning for unsupervised video summarization with diversity-representativeness reward.
\newblock In \emph{Proceedings of the AAAI conference on artificial intelligence}, volume~32, 2018.

\bibitem[Zhu et~al.(2020)Zhu, Lu, Li, and Zhou]{zhu2020dsnet}
Wencheng Zhu, Jiwen Lu, Jiahao Li, and Jie Zhou.
\newblock Dsnet: A flexible detect-to-summarize network for video summarization.
\newblock \emph{IEEE Transactions on Image Processing}, 30:\penalty0 948--962, 2020.

\end{thebibliography}

\end{document}